\documentclass[sigconf,screen]{acmart}
\usepackage{graphicx} 

\title{Search-based Testing of Vision Language Models for In-Car Scene Understanding}

\usepackage{booktabs} 
\usepackage[flushleft]{threeparttable}
\usepackage{xcolor}
\usepackage[utf8]{inputenc}
\usepackage{float}
\usepackage{ifthen}
\usepackage{caption}
\usepackage{hyperref}
\usepackage{url,moreverb,xspace}
\usepackage{enumitem}
\usepackage{array,graphicx}
\usepackage{soul}
\usepackage{balance}
\usepackage{pifont}
\usepackage{nicefrac}
\usepackage{mathtools}
\usepackage{tabularx}
\usepackage{xcolor,pifont}
\usepackage{tikz}
\usepackage{svg}
\usepackage{pdfpages}
\usepackage{array}
\usepackage{subcaption}
\usepackage[most]{tcolorbox}
\usepackage{multicol}
\usepackage{multirow}
\usepackage{pgfplots}
\usepackage{xcolor}
\usetikzlibrary{patterns}
\pgfplotsset{compat=1.18}
\usepackage{makecell}
\usepackage[normalem]{ulem}
\usepackage{lipsum}

\newcommand*\colourcheck[1]{%
  \expandafter\newcommand\csname #1check\endcsname{\textcolor{#1}{\ding{52}}}%
}
\newcommand*\colourcross[1]{%
  \expandafter\newcommand\csname #1cross\endcsname{\textcolor{#1}{\ding{56}}}%
}
\usepackage[vlined, boxruled, linesnumbered] {algorithm2e}
\SetKw{KwBy}{by}
\SetKw{KwBreak}{break}
\SetKw{KwReturn}{return}
\def\HiLi{\leavevmode\rlap{\hbox to \hsize{\color{gray!35}\leaders\hrule height .8\baselineskip depth .5ex\hfill}}}

\usepackage{amsthm}

\theoremstyle{definition} 
\newtheorem{definition}{Definition}[section] %
\newtheorem{example}{Example}[section] %

\newlength\BARWIDTH
\newlength\BARHEIGHT
\SetKwComment{Comment}{$\triangleright$\ }{}
\SetCommentSty{itshape}
\newboolean{showcomments}
\setboolean{showcomments}{true}
\ifthenelse{\boolean{showcomments}}
{\newcommand{\nb}[2] {
  \fcolorbox{black}{gray!20}{\bfseries\sffamily\scriptsize#1:}
  {\sf\small$\blacktriangleright$\textit{#2}$\blacktriangleleft$}
}
}
{\newcommand{\nb}[2]{}
}

\newcommand{\head}[1]{\noindent\textbf{#1.}}

\newcounter{fcounter}
\makeatletter
\newcommand{\thickhline}{%
    \noalign {\ifnum 0=`}\fi \hrule height 1pt
    \futurelet \reserved@a \@xhline
}
\makeatother
\usepackage{fontawesome5}

\newcommand{\toolname}{ISU-Test\xspace} %
 
\newcommand{\rs}{\textsc{RS}\xspace}

\newcommand{\gptfivechat}{\textsc{GPT-5-Chat}\xspace}

\newcommand{\geminiflash}{\textsc{Gemini-2.5-Flash}\xspace}

\usepackage{wrapfig}

\newcommand{\company}{\textsc{BMW}\xspace}

\newcommand{\gemini}{\textsc{GEMINI-2.5-Flash}\xspace}
\newcommand{\gpt}{\textsc{GPT-5-Chat}\xspace}
\newcommand{\isubmw}{\textsc{ISU\xspace}}
\newcommand{\isubmwflash}{\textsc{ISU-Flash}\xspace}
\newcommand{\moondream}{\textsc{MoonDream2}\xspace}
\newcommand{\vqa}{\textsc{VQA}\xspace}
\newcommand{\ic}{\textsc{VC}\xspace}

\usepackage{verbatim}
\usepackage{minted}

\usepackage{tablefootnote}

\usepackage[most]{tcolorbox}

\newtcolorbox{promptbox}{
  colback=gray!5,
  colframe=gray!60,
  boxrule=0.4pt,
  arc=2mm,
  left=6pt,
  right=6pt,
  top=6pt,
  bottom=6pt
}

\setcopyright{acmlicensed}
\copyrightyear{2026}
\acmYear{2026}
\acmDOI{XXXXXXX.XXXXXXX}
\acmConference[ASE '26]{Proceedings of the 41st IEEE/ACM International Conference on Automated Software Engineering}{October 12--16, 2026}{Munich, Germany}

\acmISBN{978-1-4503-XXXX-X/2026/06}

\author{Lev Sorokin}
\affiliation{
    \institution{BMW Group, Technical University of Munich}
    \city{Munich}
    \country{Germany}
}
\email{lev.sorokin@tum.de}

\author{Chen Yang}
\affiliation{
    \institution{Technical University of Munich}
    \city{Munich}
    \country{Germany}
}
\email{c.yang@gmx.net}

\author{Ken E. Friedl}
\affiliation{
    \institution{BMW Group}
    \city{Munich}
    \country{Germany}
}
\email{ken.friedl@bmw.de}

\author{Andrea Stocco}
\affiliation{
    \institution{Technical University of Munich, fortiss GmbH}
    \city{Munich}
    \country{Germany}
}
\email{andrea.stocco@tum.de}

\begin{document}
\begin{abstract}
In the automotive domain, in-car scene understanding (ISU) enables the detection of safety-critical events, such as driver distraction, and supports drivers or passengers by analyzing the in-car scene and adapting the environment (e.g., ambient lighting). The industry is increasingly exploring vision-language models (VLMs) to interpret camera-recorded in-car scenes and extract information for downstream reasoning tasks. However, VLMs may generate incomplete, erroneous, or misleading scene descriptions, highlighting the need for systematic testing. Collecting real in-vehicle data is costly, difficult to scale, and often infeasible, particularly in early design stages.
In this paper, we present \toolname, an automated testing approach that combines rendering-based scene generation with search-based testing to evaluate ISU systems. By framing testing as an optimization problem and systematically modifying scene parameters, our method generates diverse in-car scenarios and explores a wide range of configurations. We evaluate \toolname on both an industrial prototype and open-source VLMs across two case studies: question answering and captioning, comparing against randomized scenario generation. Results show that \toolname significantly outperforms the baseline, achieving up to $10\times$ higher failure rates and up to $3.6\times$ higher coverage.
\end{abstract}

\maketitle

\section{Introduction}
\label{sec:introduction}


In the automotive domain, in-cabin monitoring systems play a crucial role in the detection of safety-relevant events, such as driver distraction. Furthermore, it can be used to set up entertainment or comfort features, providing a personalized experience to control, e.g., music and light functions, or helping passengers to locate objects in the car~\cite{Diederichs2025_InCabinMonitoring, mishraIncabin2022, EuroNCAP2025_AssistedDriving}.
In addition, regulatory frameworks such as the European Union's General Safety Regulation (GSR)~\cite{EU2019_GSR}, and consumer safety assessment programs like Euro NCAP~\cite{EuroNCAP2025} increasingly mandate the deployment of driver monitoring functions, including distraction and drowsiness detection. These systems must operate reliably under a wide range of real-world conditions, making systematic testing and validation a critical requirement for certification and deployment.



A viable solution to implement in-cabin monitoring systems is to employ vision language models~\cite{Diederichs2025_InCabinMonitoring, EuroNCAP2025_AssistedDriving} to interpret 2D camera images from the vehicle's cabin and provide structured scene information for subsequent reasoning. Despite their capabilities, such models remain susceptible to providing incorrect or incomplete scene descriptions, highlighting the need for rigorous testing~\cite{humbatova2020taxonomy,RiccioEMSE20,2026-Guo-OJ-ITS}. Obtaining real-world in-cabin data is costly, difficult to scale, and often infeasible, especially during early design stages when vehicle interiors are not yet available for comprehensive data collection. While manually recorded datasets exist~\cite{sviro2024}, these are limited to specific in-cabin scenarios and have no controllability or diversity. In addition, static datasets are likely part of the training data of VLMs, limiting their usage for testing and validation practices.

\begin{figure}[t]
    \centering
    \includegraphics[width=0.95\linewidth]{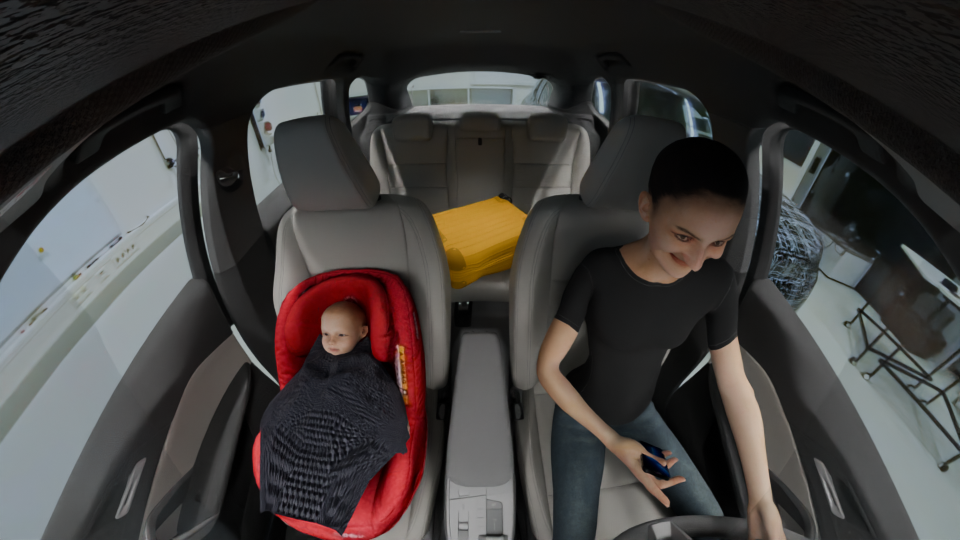}
    \caption{Rendered scene in a vehicle where a loose suitcase is placed in the back, and the driver's seat belt is not fastened.}
    \label{fig:example-scene}
\end{figure}

In this paper, we present an automated testing approach that combines controlled and rendering-based image generation with search-based testing to identify scenes where the monitoring system provides incorrect detection. This work addresses the lack of controllable and systematic validation techniques for VLM-based in-car scene understanding systems. In particular, our work focuses on the processing of single 2D images. Unlike prior work on testing DNN-based perception systems, our approach targets VLM-based scene understanding, where outputs are semantic and open-ended, requiring novel fitness and oracle definitions. 

Our approach is based on the following steps. First, we define the features on which the system needs to be tested. In our example, this includes driver features such as emotion, pose, gender, interaction with objects, static objects, and the car model. 
For passengers in particular, we employ SMPL-X~\cite{SMPL-X:2019}, which allows us to model humans parametrically according to height, weight, and pose.
For environmental parameters, such as light, we model a light source of different intensity levels and use a captured panoramic photo to define the external scene. In the second step, we parametrize the scene by (a)~altering the positions of objects, e.g., placing and positioning luggage on the back or front seats, (b)~altering the existence of objects, e.g., having the seatbelt attached or detached, and (c)~adjusting the driver's pose. The parametrization allows us to control the generation of diverse scenarios and the controlled evaluation of scenario descriptions provided by the system under test. Finally, we use search-based optimization to find failures. The optimization algorithm iteratively generates new scenes by modifying features of previous scenes guided by the evaluation results of previous executions. We evaluate our approach by testing a public and industrial prototype on the two use cases, visual question answering and captioning with multiple VLMs. The results show that \toolname can successfully and efficiently identify failures in ISU systems, outperforming random test generation approaches.

The contributions of this paper are as follows:

\head{Framework} A modular testing framework (\toolname) that combines controllable scene generation with search-based optimization to systematically expose failures in VLM-based ISU systems.

\head{Techniques} Novel fitness and oracle definitions for evaluating both structured (VQA) and open-ended (captioning) outputs of VLMs.

\head{Evaluation} An extensive empirical study on public and industrial systems demonstrating up to 10× higher failure rates and improved failure diversity compared to random testing.
\section{Background}
\label{sec:background}

In the following, we provide the definitions and illustrative examples required to understand our work. We begin with the definition of the system under evaluation and the testing problem.

\begin{definition}
An in-car scene understanding system (ISU) is a vision-language model (VLM)-based system that takes as input an image $S$, referred to as the \textit{scene}, and produces a textual representation describing relevant aspects of that scene.
\end{definition}

Depending on the interaction mode, the output can take different forms. In \textit{visual question answering} (VQA), the system generates a structured output in form of a concise answer to a given query about the scene. In \textit{visual captioning} (VC), the system produces an unstructured natural language description summarizing the scene content without a specific query.

\begin{figure}
    \centering
   \begin{promptbox}
{\footnotesize\ttfamily
You are an image analysis assistant to evaluate in-cabin scenes. Your task is to answer the following\\
questions about the visual scene and return the result in strict JSON format.\\[4pt]

\# Rules\\
- Return only the JSON.\\
- Do not provide explanations.\\
- For each question, choose exactly one answer from the given \`answer\_options\`.\\
- Do not create new answers, only choose from the provided options.\\
- If a question references something not present, select the provided ``None''.\\[6pt]

\# Questions\\[2pt]
question: "Is the driver male or female?"\\
answer\_options: ["MALE", "FEMALE"]\\[6pt]

question: "What emotion is the driver expressing?"\\
answer\_options: ["HAPPY", "SERIOUS"]\\

...
}
\end{promptbox}

    \caption{(VQA) Prompt used for a VLM in an in-car scene understanding system to describe the cabin scene.}
    \label{fig:prompt-isue}
\end{figure}

\begin{example}
Consider the scene depicted in \autoref{fig:example-scene}, where a smiling female driver is sitting next to a baby in a car seat, and a yellow luggage is visible in the backseat. An ISU can extract information such as the driver's emotions, gender, or clothing color by prompting the system in a structured manner.
    
For example, a structured output for a given prompt, such as \autoref{fig:prompt-isue}, could be: 
$R = (\text{"gender"}: \text{"female"}, \text{"seat-belt"}: \text{"off"}$, $\text{"baby\_on\_board"}: \text{"true"})$. In contrast, if the mode is visual captioning, the system is asked to provide an unstructured output to describe the scene. For instance, given the prompt \textit{``Describe what is visible in the scene, paying attention especially to humans and loose objects''}, a possible response could be: 
\textit{``The car shows a smiling female driver sitting next to a baby, holding a phone in her hand.''}  
Unstructured outputs are valuable because they allow the model to provide richer and more flexible information, including details not anticipated in a predefined schema, which can be leveraged in downstream natural language applications.
\end{example}

VLM-based systems can make inaccurate evaluations or fail in different ways to provide appropriate scene descriptions: they can wrongly classify the driver's emotions or not detect a baby in the car. Also, they could hallucinate and provide the incorrect value in the structured description or unstructured summary of the scene.
Failures can happen because of low-light conditions (dark in- and outside), partially occluded objects such as holding phones or luggage in the back, or the complexity or number of objects in the car to detect. 


    
    
    

\begin{definition}
A search-based testing problem for an ISU-System $ISU$ is defined as a tuple $P = (ISU, D, F, O)$, where

\begin{itemize}[noitemsep, topsep=0pt, leftmargin=*]

    \item $ISU$ is the in-car scene understanding system; the system under test.
    
    \item $D \subseteq \mathbb{R}^n$ is the search domain, where $n$ is the \textit{dimension} of the search space. The vector $\mathbf{s} =(x_1, \ldots, x_n) \in D$ represents the discretized representation of the scene passed to ISU. 
    
    \item $F$ is the vector-valued fitness function defined as $F: D \mapsto \mathbb{R}^m, \; F(\mathbf{x}) = (f_1(\mathbf{x}),\ldots, f_m(\mathbf{x}))$, where each $f_i$ is a scalar fitness function that assigns a quantitative score to an input based on its ability to expose faulty VLM behavior. The objective space $\mathbb{R}^m$ corresponds to the number of evaluation criteria. 
    
    \item $O$ is the oracle function, $O : \mathbb{R}^m \mapsto \{0,1\}$, which decides whether a generated scene reveals a failure. An input for which $O(F(\mathbf{x}))=1$ is considered \emph{failure-inducing}.
\end{itemize}
\end{definition}

In the following, we present our testing approach based on search-based testing to evaluate the capabilities of the ISU in providing correct scene descriptions.
\section{Approach}
\label{sec:approach}

\begin{figure*}[t]
    \centering
    \includegraphics[width=\linewidth, trim=0 4.5cm 0 2.5cm, clip]{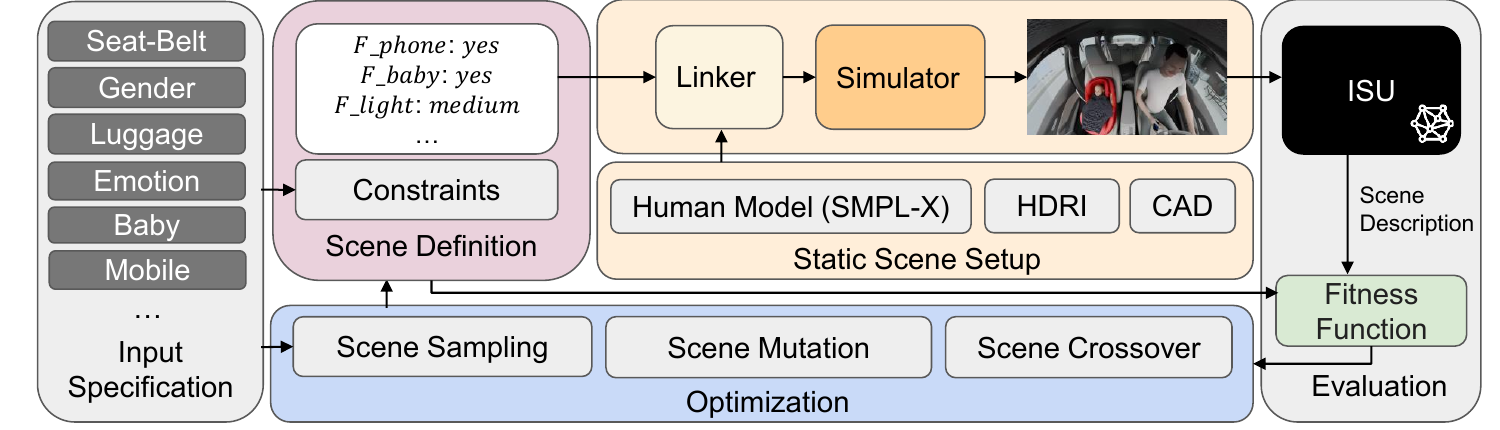}
    \caption{Overview of \toolname.}
    \label{fig:overview}
\end{figure*}

Our testing approach is illustrated in \autoref{fig:overview} and receives the following inputs: a set of features $\mathcal{F}$ describing scenes with feature domains $D_\mathcal{F}$, a fitness function $F$, an Oracle $O$, a simulator $S$, a population size $n$, and the ISU as system under test. In addition, hyperparameters, such as the testing budget, are defined.

\toolname is characterized by the following stages: Test Space Definition, Static Scene Preparation \&  Generation, Execution and Evaluation, and Optimization. 
Search-based testing is particularly suited for this setting, as the input space is high-dimensional, structured, and constrained, making exhaustive or purely random exploration inefficient. In the following, we explain each step in detail.

\subsection{Scene Specification and Representation} 

Initially a set of \textit{scene} features $\mathcal{F} = \{ \mathcal{F}_1, \mathcal{F}_2, \ldots, \mathcal{F}_n \}$ with domains $\mathcal{D_F}$  is defined to parametrize the possible space of scenes. Further $\mathcal{\hat F}$ is defined, while $\hat{\mathcal{F}} \subseteq \mathcal{F}$ is called the set of $predicted$ features. Predicted features are used in the evaluation as later explained in \autoref{sec:evaluation} to compare the system's output with the scene input.
A feature $\mathcal{F}_i$ can be numerical, categorical, or ordinal. For instance, the features for \autoref{fig:example-scene} can be defined by $\mathcal{F}_{seat}$, $\mathcal{F}_{gender}$, $\mathcal{F}_{baby}$. The domain  $\mathcal{F}_{seat}$ is categorical and for instance $D_i = \lbrace yes, no \rbrace$. A continuous feature would, for instance, model the driver's weight, while an ordinal would model the illumination ranging the values, low, medium, and high. A scene vector is defined as a set of concrete feature values $x_i \in F_i$, $F_i \in \mathcal{F}$. For instance, the scene vector for the scene in \autoref{fig:example-scene} would be $x = \lbrace belt: false, baby:yes, luggage:yes, ... \rbrace$.

\subsection{Scene Sampling} 

The first step of \toolname consists of random sampling of feature vectors. Hereby, we map all feature values based on the type, whether they are ordinal, numerical or continuous to a numerical space as proposed by Sorokin et al.~\cite{sorokin2026stellar}. I.e., all values are mapped to the range [0,1] based on min-max normalization,
Further, constraints have to be constrained to generate valid scene vectors. A possible constraint restricts, for instance, that a $luggage\_color$ is defined while no luggage exists in the scene.

\subsection{Scene Generation}

The scene generation consists of the rendering engine/simulation-based generation of the scene and the static scene setup. The static scene setup is to be performed only once, where scene elements are modeled and parametrized to be used in connection with the features defined. Hereby, we distinguish between $indoor$ assets such as vehicles cabin, luggage, bottle, baby seat, and \textit{driver-oriented assets} like the drivers pose, emotion, appearance, the \textit{adaptive assets} such as the seat-belt which should adopt to the human body shape for realistic modeling, and $environmental$-related such as the lighting in the car, external lighting or the external view around the cabin/car.

\head{Objects} 
Objects are provided through CAD models, and parametrized with respect to their position or orientation in the cabin. The granularity of the parameterization is use case specific.



\head{Driver-oriented assets} 
For human modeling, we leverage SMPL-X~\cite{SMPL-X:2019}, which is based on a data-driven parameterization of realistic human body shapes. The underlying model of SMPL-X is trained with a dataset of thousands of 3D scans of humans of different genders, poses, ages, and emotions in different setups (e.g., seating, standing). We retrieve two models for the different genders and manually adapt the models to fit in the cabin. 
The SMPL-X technique allows us to automatically generate different body shapes of the driver. To model the emotional state, we tune the coefficient of several shapekeys on the SMPL-X human’s face. To model the clothing of the driver, we manually provide different overlays based on the parametrization of different overlays in the desired color and type of clothing.

\head{Adaptive assets} For adaptive modeling, such as seat belt modeling, we employ surface modeling and apply a shrink-wrap modifier~\cite{blender_shrinkwrap_36} to position the seat belt over the passenger’s body while avoiding intersections with the body mesh.

\head{Environmental assets} Environmental lights are modeled by providing functions that set the internal state of the light source in the simulator scene, where the lighting intensities depend on the simulator's rendering engine. For the external view, we record first an image of the exterior of the car in the selected environment in the HDRI format.

\head{Novel Scene Synthesis} 
For a given scene vector, all feature values are forwarded based on the corresponding feature type to the respective function to set up the scene component based on the linking from the previous step. When all features have been applied, the rendering of the image is triggered, producing an image of the in-cabin scene.

\head{Fitness Function Definition}\label{sec:fitness-functions}
For the fitness function, we distinguish between \textit{response-related} and diversity-optimization related fitness functions. 

The \textbf{response-oriented} fitness function assesses the quality of the response and receives as input the set of prediction features $\mathcal{\hat F}$ and the system response $R$ and outputs a score between 0 and 1, where 0 is the worst score, and 1 is the best score. The implementation of the fitness function depends on the execution mode of ISU. 

\textit{1) Question Answering.} For question answering, the fitness function compares the values of predicted features in the input vs. the values of those in the output for exact matches, i.e., it is defined  as follows:


\[
F_{\mathrm{qa}} = 
\frac{1}{|\mathcal{C}|}
\sum_{F_i \in \mathcal{C}}
w_{F_i} \cdot \mathbf{1}\{\hat{f}_i = f_i\}.
\]

where $C$ is the set of features, and $w_{F_i}$ is a so-called feature weight defining the importance of the detection of the corresponding feature in the scene. Weighting allows us to penalize incorrect outputs for features that are more important than others, e.g., those related to the driver or a child in the car.

\textit{2) Visual Captioning.} For captioning evaluation, we first generate a reference caption by instantiating a parametrized text template with the scene vector $X$. An example template is \textit{The scene shows a \textbf{gender} driver with $\textbf{t-shirt color}$ t-shirt having the emotion $\textbf{emotion}$}. The evaluation of the output scene description is performed by evaluating the similarity between the reference caption and the generated caption, i.e., we apply the following four metrics to capture different evaluation dimensions and assess the generated scene description:

\textit{Embedding Similarity ($F_{\mathrm{emb}}$)}: The embedding similarity captures the description semantics on the sentence level. We embed the reference and predicted captions using a text encoder, i.e., \\ \texttt{all-MiniLM-L6-v2}, and compute cosine similarity between the encodings.

\textit{BLEU Score ($F_{\mathrm{bleu}}$}): BLEU measures the $n$-gram overlap between generated and reference text and allows us to assess the lexical alignment and local phrase correctness~\cite{bleu}.

\textit{METEOR score ($F_{\mathrm{meteor}}$}): METEOR focuses on the precision, recall, and word order with stemming and synonym, balancing strict overlap metrics such as BLEU~\cite{meteor}.

\textit{BERTscore ($F_{\mathrm{bert}})$:} BERTScore provides contextual semantic alignment at the token level. Unlike sentence-level embeddings, BERTScore computes semantic similarity via contextual token embeddings and cosine similarity~\cite{zhang2019bertscore}.

The \textbf{diversity-oriented} fitness function $F_{div}$ evaluates the diversity of a scene compared to previously generated scenes by measuring for non-continuous features, its Hamming Distance to other scenes~\cite{hamming1950error}, and for continuous features, the actual normalized Euclidean distance. In particular, for two scenes \( s \) and \( s' \) the function is defined as

\begin{equation}
F_{\mathrm{diversity}}(s, s') =
\sum_{i \in \mathcal{D}} \mathbf{1}\!\left[f_i \neq f_i' \right]
\;+\;
\sum_{i \in \mathcal{C}} \frac{\lvert f_i - f_i' \rvert}{\Delta_i},
\end{equation}

where $\mathcal{D}$ are discrete and $\mathcal{C}$ continuous features.

\head{Oracle Function} The oracle function defines when a test is failing and is defined in a use case specific. For the question answering-based mode, it is defined as:

    \begin{align*}
        O(test) : F_{qa} \leq th
    \end{align*}

while for the captioning-based mode it is defined as:

    \begin{align*}
        O : (F_{\mathrm{emb}} \leq th_{emb}) \wedge (F_{\mathrm{bleu}} \leq th_{bleu}) \wedge \\ (F_{\mathrm{meteor}} \leq th_{meteor}) \wedge (F_{\mathrm{bert}} \leq th_{bert})
    \end{align*}

where $th$, $th_{emb}$, $th_{bleu}$, $th_{meteor}$ and $th_{bert}$ are thresholds. 

\head{Optimization Algorithm} All defined fitness functions are used by the optimization algorithm to rank executed tests based on a score to guide the generation of new scenes. While the response-oriented fitness function/s is/are to be minimized to prioritize tests that fail for guiding the generation of new scenes, the diversity fitness function is to be maximized to foster detection of diverse scenes.

In our study, we employ a genetic population-based algorithm to guide the generation of failure-revealing scenes. In particular, the algorithm applies genetic operators such as mutation or crossover to generate promising scenes.

\head{Scene Mutation and Crossover} The mutation and crossover operators allow the generation of new scenes from existing scenes. While the mutation operator modifies single feature values, e.g., changing the seat belt from on to off, or lighting medium to low, the crossover operator recombines features of two scenes to generate two new scenes. The recombination of two scenes can lead to scenes where features become incompatible. We impose, therefore, scene constraints after generation, such as those applied after scene sampling and selecting predefined feature values without conflicts. An example restriction would be placing both luggage and a child in the same place. If a scene vector is generated, which has already been witnessed before, a new scene is randomly sampled (duplicate elimination).

\head{Survival} In this step, a fixed number of scene vectors/candidates is selected based on the fitness values to be taken over to the next iteration of the algorithm. This step is necessary as otherwise the set of test cases to be executed would always grow, and candidates would remain in the population with poor fitness scores.

When the testing budget is exhausted, \toolname invokes the oracle to classify the generated conversations based on their fitness values and outputs the set of failing test cases during the search.
\section{Evaluation}
\label{sec:evaluation}

Our evaluation considers the following research questions:

\subsection{Research Questions}

\head{RQ\textsubscript{1} (effectiveness)} \textit{How effective is \toolname in identifying failures?}

\head{RQ\textsubscript{2} (diversity)} \textit{How diverse are failures identified by \toolname?}

\head{RQ\textsubscript{3} (validation)} \textit{What is the validity rate of identified failures when comparing rendered scenes of found failures to real-world scenes?}


In the first place, we want to evaluate with RQ\textsubscript{1} how effective our approach is in finding failures, which is the primary goal when performing automated testing.

However, the failing scenes identified may be highly similar, for instance, when scenes differ only by a minimal change in the orientation of a piece of luggage. Prior research has emphasized~\cite{surrogate2024biagiola, feldt2016diversity} the importance of identifying diverse failures to support debugging and fault localization. Therefore, in RQ\textsubscript{2}, we evaluate the diversity of failures detected by \toolname.

In RQ\textsubscript{3}, we aim to evaluate the extent to which the system under test produces consistent description outcomes when provided with a reconstructed real-world scene in place of a simulated scene.

\subsection{Study Subjects}

We evaluate our approach on the case studies of visual question answering (VQA) and visual captioning (VC).
In VQA, a scene is provided to the system with the goal of retrieving a structured description of the scene with a predefined format (features predefined). In visual captioning, the system under test has to generate a caption summarizing the scene.

For VQA, we evaluate three standalone general-purpose VLM models from three different providers, namely \gptfivechat, \geminiflash, and \moondream, a small-sized VLM of size 2B, and two industrial prototypes from our partner company \company. The industrial systems are available in two variants, $ISU-Flash$ and $ISU$, where in the first version a more cost-efficient and lightweight node is used than in the second variant. Both variants natively output a structured JSON dictionary in the required format. 

For general-purpose VLM models, we instruct the model with the prompt as shown in Figure 5.1. The complete prompt can be found in our supplementary material~\cite{repo}. The selection of the models is guided by the different capabilities, pricing, size, and inference latency of the models. For the industrial system, we adopt the prompts to capture all features modeled in our scenes.

\autoref{tab:isu_suts} provides an overview of all evaluated configurations. 


\begin{table}[t]
    \centering
        \caption{List of test subjects used in our study.}
    \label{tab:isu_suts}
    \begin{tabular}{llll}
        \toprule
        \textbf{Category} & \textbf{Model} & \textbf{Version} & \textbf{Case Study} \\
        \midrule
        OpenAI     & \gpt         & 2025-04-01  &  VQA, VC \\
        Google     & \gemini      & via VertexAI       &  VQA, VC \\
        OpenSource & \moondream   & 2025-06-21         &  VQA, VC \\
        Industrial & \isubmw      & 2025               &  VQA \\
        Industrial & \isubmwflash & 2025               &  VQA \\
        \bottomrule
    \end{tabular}
\end{table}

\begin{table}[t]
\centering
\caption{Overview of features used in our study.}
\label{tab:case_study_features}
\setlength{\tabcolsep}{3pt}
\small
\begin{tabularx}{\columnwidth}{l l X}
\toprule
\textbf{Cat.} & \textbf{Feature} & \textbf{Values} \\
\midrule

\multirow{7}{*}{Driver}
& Gender & Female, Male \\
& Emotion & Happy, Serious \\
& Phone Pos. & None, P1–P10 \\
& Weight & W1–W4 \\
& Height & H1–H4 \\
& Seatbelt & Y/N \\
& T-shirt & Black, White \\

\midrule

\multirow{3}{*}{Child}
& Seat Exists & Y/N \\
& Child in Seat & Y/N \\
& Orientation & Front, Back \\

\midrule

\multirow{2}{*}{Env.}
& Car Lights & On/Off \\
& External Light & High, Medium, Low \\

\midrule

\multirow{7}{*}{Objects}
& Luggage Exists & Y/N \\
& Orientation & 20–140 \\
& Position & Passenger, Back \\
& Colour & Yellow, Anthracite \\
& Phone Exists & Y/N \\
& Coke Exists & Y/N \\
& Can Exists & Y/N \\

\bottomrule
\end{tabularx}
\end{table}

\begin{table}[t]
\centering
\caption{Feature weights used in the VQA case study.}
\label{tab:feature_weight}
\begin{tabular}{p{7cm} c}
\toprule
\textbf{Features} & \textbf{Weight} \\
\midrule
Gender, Emotion, Cola Bottle, Cola Can & 0.05 \\
Phone, Seat Belt, Luggage Existence, Luggage Location, & \\
Codriver Phone, Child Seat, Child Seat Orientation, Child Existence & 0.10 \\
\bottomrule
\end{tabular}
\end{table}

\begin{table*}[t]
\centering
\caption{Failure rate (\%, mean $\pm$ standard deviation) for \vqa and \ic across SUTs under \toolname and \rs.}
\label{tab:failure_cases_merged}
\setlength{\tabcolsep}{4pt}
\resizebox{\textwidth}{!}{
\begin{tabular}{l cc cc cc cc cc}
\toprule
\textbf{Thr.}
& \multicolumn{2}{c}{\textbf{\gpt}}
& \multicolumn{2}{c}{\textbf{\gemini}}
& \multicolumn{2}{c}{\textbf{\moondream}}
& \multicolumn{2}{c}{\textbf{\isubmw}}
& \multicolumn{2}{c}{\textbf{\isubmwflash}} \\
\cmidrule(lr){2-3}\cmidrule(lr){4-5}\cmidrule(lr){6-7}
\cmidrule(lr){8-9}\cmidrule(lr){10-11}
& \toolname & \rs
& \toolname & \rs
& \toolname & \rs
& \toolname & \rs
& \toolname & \rs \\
\midrule

\multicolumn{11}{c}{\textbf{\vqa}} \\
\midrule

\textbf{Tests}
& 606.7$\pm$10.3 & 560.3$\pm$7.9
& 563.3$\pm$15.1 & 538.2$\pm$10.2
& 286.7$\pm$10.3 & 200.0$\pm$16.0
& 490.0$\pm$11.0 & 477.7$\pm$6.1
& 606.7$\pm$30.1 & 570.3$\pm$29.1 \\

1.0
& \textbf{88.2}$\pm$7.5 & 71.7$\pm$0.8
& \textbf{60.0}$\pm$8.4 & 38.2$\pm$2.5
& 100.0$\pm$0.0 & 100.0$\pm$0.0
& \textbf{66.5}$\pm$8.5 & 37.2$\pm$2.4
& \textbf{61.7}$\pm$7.6 & 30.0$\pm$1.1 \\

0.9
& \textbf{46.5}$\pm$7.2 & 18.7$\pm$1.2
& \textbf{29.7}$\pm$2.6 & 4.2$\pm$0.8
& 100.0$\pm$0.0 & 100.0$\pm$0.0
& \textbf{32.7}$\pm$6.1 & 12.2$\pm$1.7
& \textbf{23.2}$\pm$3.4 & 6.3$\pm$1.2 \\

0.8
& \textbf{10.5}$\pm$3.6 & 1.8$\pm$0.4
& \textbf{15.2}$\pm$2.9 & 1.5$\pm$0.5
& 100.0$\pm$0.0 & 100.0$\pm$0.0
& \textbf{20.8}$\pm$5.2 & 5.7$\pm$0.5
& \textbf{14.2}$\pm$2.5 & 2.3$\pm$0.5 \\

0.7
& \textbf{3.8}$\pm$2.1 & 0.0$\pm$0.0
& \textbf{1.7}$\pm$2.0 & 0.0$\pm$0.0
& \textbf{98.0}$\pm$1.5 & 95.5$\pm$1.9
& \textbf{5.0}$\pm$2.8 & 0.2$\pm$0.4
& \textbf{2.5}$\pm$1.5 & 0.0$\pm$0.0 \\

0.6
& 0.0$\pm$0.0 & 0.0$\pm$0.0
& \textbf{0.5}$\pm$0.5 & 0.0$\pm$0.0
& \textbf{90.2}$\pm$5.4 & 80.5$\pm$8.3
& 0.0$\pm$0.0 & 0.0$\pm$0.0
& 0.0$\pm$0.0 & 0.0$\pm$0.0 \\

0.5
& 0.0$\pm$0.0 & 0.0$\pm$0.0
& \textbf{0.2}$\pm$0.4 & 0.0$\pm$0.0
& \textbf{64.0}$\pm$7.2 & 52.5$\pm$20.0
& 0.0$\pm$0.0 & 0.0$\pm$0.0
& 0.0$\pm$0.0 & 0.0$\pm$0.0 \\

\midrule

\multicolumn{11}{c}{\textbf{\ic}} \\
\midrule

\textbf{Tests}
& \textbf{180.0}$\pm$6.3 & 124.2$\pm$12.2
& \textbf{166.7}$\pm$10.3 & 113.3$\pm$17.1
& \textbf{156.7}$\pm$5.2 & 88.8$\pm$3.1
& -- & --
& -- & -- \\

0.8
& \textbf{99.7}$\pm$0.8 & 99.2$\pm$0.8
& \textbf{96.0}$\pm$2.0 & 87.5$\pm$4.2
& 100.0$\pm$0.0 & 100.0$\pm$0.0
& -- & --
& -- & -- \\

0.7
& \textbf{88.0}$\pm$6.2 & 72.3$\pm$4.6
& \textbf{58.3}$\pm$6.2 & 27.5$\pm$4.5
& \textbf{96.2}$\pm$4.2 & 92.3$\pm$8.7
& -- & --
& -- & -- \\

0.6
& \textbf{36.3}$\pm$12.1 & 13.8$\pm$3.5
& \textbf{19.3}$\pm$4.5 & 4.0$\pm$2.1
& 53.7$\pm$25.2 & \textbf{54.5}$\pm$39.4
& -- & --
& -- & -- \\

0.5
& \textbf{3.5}$\pm$2.9 & 0.5$\pm$0.5
& \textbf{2.2}$\pm$2.6 & 0.0$\pm$0.0
& 12.8$\pm$14.4 & \textbf{13.3}$\pm$14.8
& -- & --
& -- & -- \\

\bottomrule
\end{tabular}}
\end{table*}

\subsection{Metrics}

\head{RQ\textsubscript{1}} We evaluate the effectiveness of \toolname by measuring the number of failing scenarios identified and the ratio between failures and the number of scenes produced.

\head{RQ\textsubscript{2}} To evaluate the diversity of found failing scenes, we first map scenes applying CLIP embeddings into a numerical space from all runs and cluster the mapped scenes. Then we apply k-means-based clustering to retrieve a finite set of failure clusters. Then, for each approach, we evaluate coverage of the approximated clusters as well as the failure distribution among the clusters as proposed by Biagiola et al.~\cite{surrogate2024biagiola}.

\head{RQ\textsubscript{3}} To evaluate how well the generated failing test inputs correspond to actual failing tests, we first sample for each testing approach over all runs and recreate a portion of the scenarios in reality. In the first step, both the rendered images and the recreated real-world scenes are presented to human evaluators, who are asked to assess whether they depict the same scenario. In the second step, the performance of the system under test is evaluated using both the simulated images and the recreated scenes.

\subsection{Experimental Configuration \vqa}

For VQA, we select the features as defined in \autoref{tab:case_study_features}. The features and their parametrization have been first derived based on studies from relevant use cases available in the literature~\cite{driveandact2019,Diederichs2025_InCabinMonitoring} and then confirmed and refined with experts from the industrial partner.

\subsubsection{Fitness Function}
\label{isu_fitness}

The fitness evaluation for \vqa consists of two fitness functions: $F_{\mathrm{cat}}$, and $F_{\mathrm{diversity}}$. While the first is minimized to find failing tests, the second is to be maximized to find more diverse scenes.
The weighting of features is illustrated in \autoref{tab:feature_weight} and has been defined based on discussions with \company experts. The weighting can be adopted based on the use cases and is chosen in a way to penalize safety-related features more than non-safety-related (like finding small objects in a car).

\subsubsection{Oracle Function}
\label{sec:vqa_oracle}

For \vqa, a test is classified as failing if at least one feature classification is incorrect, i.e., $O: F_{\mathrm{qa}} < 1.0$. We additionally report results under relaxed thresholds to study the failure severity when considering the custom weighting of features.



\subsection{Experimental Configuration VC}

\ic is a canonical open-ended VLM task \cite{karpathy2015deepvisualsemanticalignmentsgenerating,vinyals2015tellneuralimagecaption} and complements the \vqa case study~\cite{agrawal2016vqavisualquestionanswering}. Unlike \vqa, \ic does not admit a single correct answer~\cite{bernardi2017automaticdescriptiongenerationimages}, which makes oracle construction inherently difficult. Correctness is therefore typically assessed using similarity-based metrics rather than exact matching~\cite{vedantam2015ciderconsensusbasedimagedescription,kilickaya2016reevaluatingautomaticmetricsimage}. We use this case study primarily to study the robustness of our approach when using different processing techniques in the SUT.
We use the same search budget, the same hyperparameter configuration, and the same features as defined for \vqa.

\subsection{Fitness}

To evaluate the fitness, we use four functions as defined in \autoref{sec:fitness-functions}, including the diversity-oriented function.
We provide a predefined template which is filled as explained in \autoref{sec:fitness-functions} with values of the corresponding feature vector to generate a reference caption. An example reference prompt for the scene vector is provided in \autoref{fig:gt_captioning}.

\begin{figure}[htbp]
    \centering
   \begin{promptbox}
{\footnotesize\ttfamily
The image is taken from inside a car parked in a \textcolor{blue}{garage}. A \textcolor{blue}{happy} \textcolor{blue}{male} driver wearing a \textcolor{blue}{black} T-shirt is seated in the car. The driver is \textcolor{blue}{not wearing a safety belt}. A \textcolor{blue}{anthracite} suitcase is visible on \textcolor{blue}{rear seat}. A \textcolor{blue}{rear-facing} baby seat \textcolor{blue}{with a baby} is positioned on the front passenger seat.
}
\end{promptbox}
    \caption{Caption derived from ground-truth parameters. Blue text segments are injected from the scene vector.}
    \label{fig:gt_captioning}
\end{figure}

\subsection{Oracle}

For the oracle, we only apply a threshold for the first fitness function $F_{llm}$ and use a threshold of 0.65. We select the threshold value empirically based on preliminary experiments.

\subsection{Results Effectiveness (RQ\textsubscript{1})}

\begin{table*}[t]
\centering
\caption{Diversity analysis under CLIP encoding for \vqa and \ic.
Each cell reports Coverage (\%) and Entropy.}
\label{tab:diversity_combined}

\setlength{\tabcolsep}{5pt}
\resizebox{\textwidth}{!}{
\begin{tabular}{l cc cc cc cc cc}
\toprule
\textbf{Alg.}
& \multicolumn{2}{c}{\gpt}
& \multicolumn{2}{c}{\gemini}
& \multicolumn{2}{c}{\moondream}
& \multicolumn{2}{c}{\isubmw}
& \multicolumn{2}{c}{\isubmwflash} \\

\cmidrule(lr){2-3}\cmidrule(lr){4-5}\cmidrule(lr){6-7}
\cmidrule(lr){8-9}\cmidrule(lr){10-11}

& Coverage & Entropy
& Coverage & Entropy
& Coverage & Entropy
& Coverage & Entropy
& Coverage & Entropy \\

\midrule

\multicolumn{11}{c}{\textbf{\vqa}} \\
\midrule

\toolname
& 100.00 & \textbf{97.48}
& 100.00 & \textbf{92.84}
& 100.00 & 89.18
& \textbf{96.25} & \textbf{94.80}
& \textbf{97.52} & 94.49 \\

\rs
& 100.00 & 91.52
& 100.00 & 91.37
& 100.00 & \textbf{91.74}
& 96.18 & 94.44
& 95.37 & 94.46 \\

\midrule

\multicolumn{11}{c}{\textbf{\ic}} \\
\midrule

\toolname
& \textbf{82.61} & \textbf{92.52}
& \textbf{86.35} & \textbf{91.61}
& \textbf{79.31} & \textbf{91.67}
& -- & --
& -- & -- \\

\rs
& 37.83 & 80.42
& 23.74 & 68.52
& 73.64 & 92.05
& -- & --
& -- & -- \\

\bottomrule
\end{tabular}}
\end{table*}

The effectiveness results comparing \toolname with random search (RS) are shown in \autoref{tab:failure_cases_merged}. We can see that, for both case studies, \toolname achieves for the majority of compared systems and threshold higher scores than the randomized baseline approach. 

In particular, for the \vqa cases study \toolname achieves a threshold of 1 to 0.7 for all comparisons, with higher failure rates for the open-source as well as the industrial case study. A less strict oracle with a lower threshold below 0.7 allows only for the more efficient but smaller models \geminiflash and \moondream to detect failures. Also, for this configuration, setups \toolname still outperform the random baseline in the failure discovery. However, the highest failure rates are found in \gptfivechat and \moondream for the most strict threshold of 1.0.

For the \ic study, we can observe that for thresholds between 0.9 and 1, there is no difference between the approaches, exhibiting 100\% failure rates. This strict comparison rather considers expression variations and likely does not take semantic variations into account, making it not suitable for comparison of the systems' scene evaluation behaviour.
For the thresholds 0.8 to 0.5, we can see that \toolname outputperforms the randomized baseline for the majority of comparisons (11 out of 12).

In addition, we have performed the statistical Wilcoxon test~\cite{Wilcoxon1945} and evaluated the Vargha-Delaney effect size~\cite{Vargha-Delaney} to assess whether the differences are statistically significant. The results show that all differences are statistically significant with large effect sizes. Further results showing the number of failures over time can be found in the replication package~\cite{repo}.

\begin{tcolorbox}[boxrule=0pt,frame hidden,sharp corners,enhanced,
borderline north={1pt}{0pt}{black},borderline south={1pt}{0pt}{black},
boxsep=2pt,left=2pt,right=2pt,top=2.5pt,bottom=2pt]
\textbf{RQ\textsubscript{1} (Effectiveness).}
\toolname consistently identifies a substantially larger number of failures and achieves higher failure rates than random testing across most configurations in both case studies. In visual question answering, the greatest improvement is observed with a Gemini-based VLM, showing up to a 10× higher failure rate. For captioning, the same system exhibits the largest gain, with failure rates up to 5× higher than those obtained through random testing.
\end{tcolorbox}

\subsection{Results Diversity (RQ\textsubscript{2})}

The diversity results are presented in \autoref{tab:diversity_combined}, which shows both the coverage of failure clusters and the entropy. We can see that for \vqa, the coverage for \toolname and \rs are maximum and equally high for the open source systems, while for the industrial system, the values are slightly higher for \toolname. Regarding entropy, we see that for four out of five comparison values for \toolname are higher.
For visual captioning, we can see that both coverage and entropy are always higher for \toolname than for \rs, reaching up to $3.6\times$ higher coverage and 34\% higher entropy than \rs.

The statistical test results show that the difference in entropy is for both case studies, and that for \ic also the coverage difference is statistically higher with large effect sizes.

\begin{tcolorbox}[boxrule=0pt,frame hidden,sharp corners,enhanced,
borderline north={1pt}{0pt}{black},borderline south={1pt}{0pt}{black},
boxsep=2pt,left=2pt,right=2pt,top=2.5pt,bottom=2pt]
\textbf{RQ\textsubscript{2} (Diversity).}
\toolname consistently achieves higher entropy with a maximal increase of 34\%, indicating a more balanced distribution of failures across failure clusters in both case studies. At the same time, coverage values for \vqa are maintained at levels comparable to randomized testing. For \ic, optimization-based testing further improves coverage up to 3.6$\times$ compared to randomized testing.
\end{tcolorbox}

\begin{figure*}[t]
  \centering
    \includegraphics[width=0.9\textwidth]{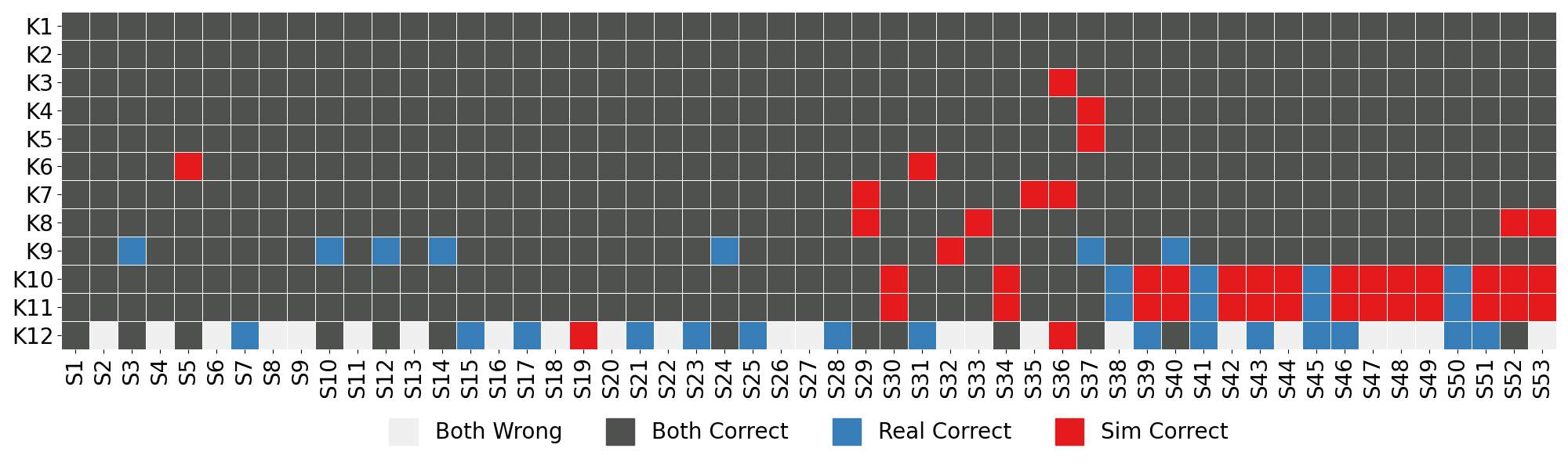}
    \caption{
    Agreement between real and simulated evaluations.
    Features: K1 phone\_codriver\_seat, K2 gender, K3 baby\_seat, K4 colacan\_codriver\_seat,
    K5 colabottle\_codriver\_seat, K6 emotion, K7 baby, K8 phone\_driver,
    K9 safety\_belt, K10 suitcase\_location, K11 suitcase, K12 baby\_seat\_orientation.
    Samples are indexed by S1--S53.
    }
    \label{fig:agreement_heatmap_real_vs_sim}
\end{figure*}

\subsection{Results Validity (RQ\textsubscript{3})}\

To study the agreement between test executions with rendering-based generated scenes from \toolname and scenes in real life, we recreate a portion of rendered scenes in a real vehicle and curate a dataset of 53 image pairs. The dataset is sampled randomly from both failing and passing tests generated during runs of \isubmwflash. Data collection involves three participants with different but feature-consistent body characteristics modeled in \toolname regarding weight and height, including two males and one female. Static assets such as the mobile phone, cola can, door keys, and baby seat are provided as real objects. For the baby, a baby doll is used. The total recreation process requires up to 6 hours because of the number of features and the type of physical assets varied.

An example recreation is shown in \autoref{fig:sim-augment-real}.
In the first step, we assess whether the reconstructed/real scenes correspond \textit{semantically} to the simulated scenes. For this, we employ three human annotators who were not involved in the development of \toolname and decide on the alignment for each scene based on majority voting. The evaluation shows a reconstruction validity rate of 100\%.

In the second step, to evaluate the system's behaviour when using real scenes, we pass both the real and reconstructed scenes to  \isubmwflash. We select here \isubmwflash as it provides lower failure rate scores in the previous analysis compared to \isubmw. To take into account the stochastic nature of \isubmwflash, we pass each scene three times and select the majority of the predicted labels per feature. Outcomes are categorized into four cases: (i) both predictions correct, (ii) both incorrect, (iii) correct only for the synthetic image, and (iv) correct only for the real image (see \autoref{fig:agreement_heatmap_real_vs_sim}). The agreement rate is then defined as:

\[
\text{Agreement Rate}
= \frac{\text{Sim/Real Correct} + \text{Sim/Real Wrong}}{\text{All Features Compared}}.
\]

The evaluations show an agreement rate of 89\%.
We observe that especially driver-related features such as \emph{emotion detection}, \emph{phone existence}, and \emph{gender} are robust across both domains (disagreement 2\%), while the scenes including luggage in the back or when the driver is belted wearing a black shirt are often misclassified. This could be attributed to the fact that, on the one hand, luggage is partially obscured by the front seats, making the correct recognition difficult. On the other hand, a black belt in simulation can become overlooked because of a similar color appearance.

\begin{tcolorbox}[boxrule=0pt,frame hidden,sharp corners,enhanced,
borderline north={1pt}{0pt}{black},borderline south={1pt}{0pt}{black},
boxsep=2pt,left=2pt,right=2pt,top=2.5pt,bottom=2pt]
\textbf{RQ\textsubscript{3} (Validity).}
\toolname generated images show a validity rate of 89\% when comparing the system performance on simulated vs. recreated scenes.
\end{tcolorbox}

\head{Key Insights}
Our results reveal three main insights. First, optimization-based testing is substantially more effective than random exploration in identifying failure-inducing scenarios, particularly under strict correctness thresholds. Second, while failure coverage for VQA saturates quickly, entropy analysis shows that \toolname explores a more balanced and diverse set of failure modes. Third, the high agreement between simulated and real-world evaluations suggests that simulation-based testing can serve as a reliable proxy for early-stage validation of VLM-based systems.


\subsection{Threats to Validity}\label{sec:threats}


\head{Internal Validity} To assess the validity of oracle results, we evaluated the non-determinism of the system under test by repeating identical inputs and measuring output consistency. The system achieved over 90\% consistency across runs, indicating limited variability. Therefore, no additional mechanisms to control non-determinism were applied during the search process.

\head{Construct Validity}
Human–object interactions (e.g., a driver holding a mobile phone) are modeled using a simplified animation-based approach instead of a physics-based model, prioritizing controllability over realism. This may limit interaction fidelity and pose a threat to construct validity, particularly for tasks requiring fine-grained physical cues.

\head{External Validity} The approach is evaluated on in-car scene understanding for a single car model. It can generalize to other domains and models whose scenes are composed of assets that can be represented in 3D and rendered using engines such as Maya or Blender—for example, activity monitoring in enclosed environments or the aircraft domain. Fitness and oracle functions need to be defined for each specific use case; however, the overall framework (e.g., weighted averages over prediction matches or threshold-based oracles) can remain unchanged.

The light exposure alignment in our study is tailored to a specific in-car camera and replicates its adaptive exposure behavior in the rendering pipeline. As exposure and calibration are camera-dependent, transferring the approach to other sensor setups may require additional adaptation.
Regarding the scene parametrization, feature values have been selected in a way to be able to explore diverse scenes, balancing the possibility for recreation and failure validation.
Feature values can span different and bigger search intervals for other applications.
A further limitation concerns the abstraction of the feature space, which may not fully capture complex dependencies between features (e.g., correlated driver behaviors and environmental conditions), potentially limiting realism in certain edge cases.
This suggests that the approach is applicable to other domains involving structured visual environments, such as robotics, surveillance, or human–machine interaction systems.


\section{Qualitative Analysis}\label{sec:discussion}

\subsection{Improving Transferability to Real World} 
To study how data transformation techniques could further improve the quality of the generated scenes, we applied data transformation as a postprocessing step, specifically employing Neural Style Transfer (NST) on the simulated images~\cite{gatys2016nst}. In particular, we used 300 simulated images from prior executions, together with a single reconstructed scene, to fine-tune an existing VGG model~\cite{gatys2016nst}. We manually evaluated different model configurations trained with different hyperparameters responsible for the style and content transfer, and used default values for early stopping and max epoch size. We manually selected the hyperparameter setup, yielding the best result in terms of stylization of the image. Example results are provided in \autoref{fig:sim-augment-real}. 

\begin{figure*}[h]
    \centering
    \includegraphics[width=1\linewidth]{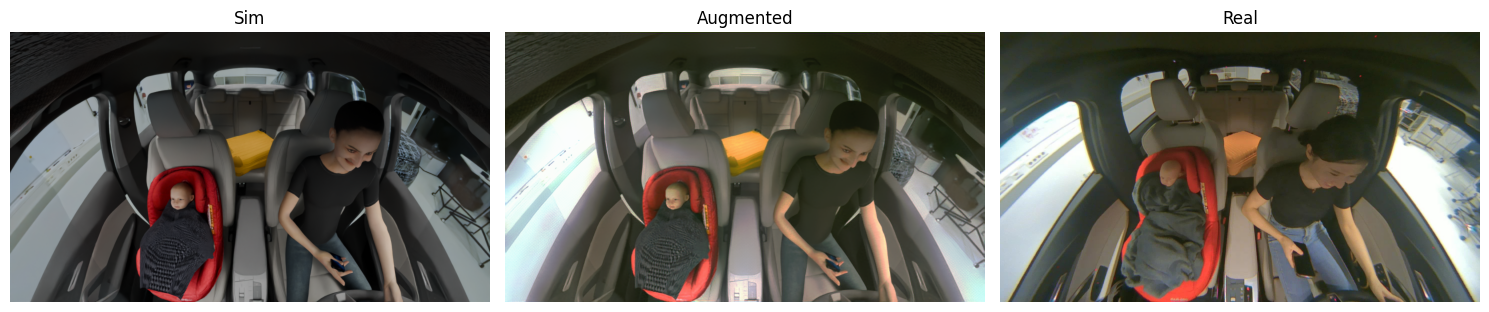}
    \label{fig:sim-augment-real-2}
\end{figure*}
\begin{figure*}
    \centering
    \includegraphics[width=1\linewidth]{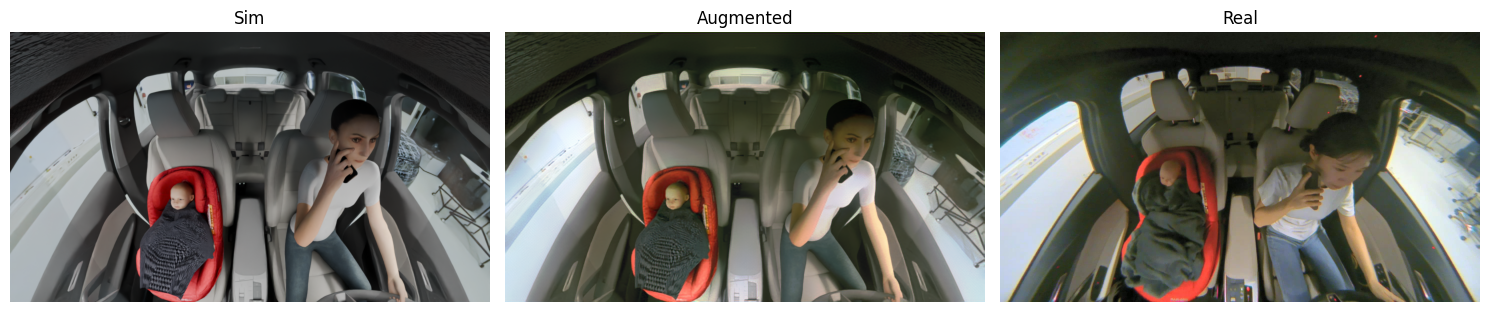}
    \caption{Overview of two different scenes (top/bottom): rendered scene (left), augmented scene using neural style transfer (middle), and reconstructed scene (right).}
    \label{fig:sim-augment-real}
\end{figure*}
Our results show that augmentation increases the agreement rate by 2.6\% over all feature classifications. We can observe that in general, the coloring and the blurriness appear for augmented images closer to the recreated images.
The reason for the improved rate could be that objects or features that are not detectable because of their appearance in reality also become undetectable after transformation.
To assess whether NST processed images are closer to real-world images, we evaluate the distributional similarity using the Fréchet Inception Distance (FID) and Kernel Inception Distance (KID)~\cite{2024-Lambertenghi-ICST}. 
 The results show that NST reduces the FID distance by 7.6 scores compared to only simulated images (FID with NST: 151.9 vs. FID for simulated 159.5), while for KID the improvement is 0.025 (KID with NST: 0.15 vs. 0.175) with variance below 0.007. While the results indicate that NST improves the realism of simulated scenes, absolute distance values remain high.

\head{Other Scene Transformation/Augmentation Experiments} 
In principle, beyond NST, vision–language model–based approaches such as NanoBanana~\cite{nanobana2026} and \texttt{GPT-Image-1}, as well as ControlNet-based methods (e.g., NVIDIA Cosmos)~\cite{nvidiacosmos2026}, could be integrated into \toolname to generate more photorealistic scenes. However, preliminary experiments indicate that these approaches do not preserve semantic consistency. For example, the position of the driver's arms may change, or the camera viewpoint may shift (e.g., from a frontal to a lateral view). Such inconsistencies adversely affect failure assessment and limit their suitability for integration into \toolname. Similarly, image-to-image-based techniques can be applied to modify clothing colours or allow a visually more realistic seat-belt modeling. Applicable methods include in-painting~\cite{baresi2025}, diffusion-based augmentation~\cite{baresi2025}, and ControlNet-based approaches~\cite{alimisis2025advances}.
However, as before, the scene geometry and semantics were not preserved, which is why further research is required to understand how such techniques can be integrated in a robust manner.

\subsection{Failure analysis} We analyzed the failures for both case studies for \vqa and \ic to understand the different potentials of the underlying VLMs/SUTs. For \vqa we used feature-wise comparisons, while \ic requires manual inspection as the output is unstructured. We excluded the model \moondream from the analysis because of its high failure rates.

For visual question answering, across both test approaches \rs and \toolname, for the systems \geminiflash,\isubmw, and \isubmwflash the most frequent misclassifications happen for \emph{baby seat orientation} with up to 45\%, followed by suitcase misdetection with 23.5\% for \toolname, while for \rs the second highest failure category is still baby seat orientation but in the inverse direction.

\gptfivechat fails most on emotion detection, detecting a happy person as serious (43\%, vs. 48\%), followed similarly by a baby seat direction misclassification with 27\% vs 21\% failure rates. The complete failure distribution can be found in our replication package. Reviewing samples of executed tests in the captioning case study shows that generated tests often contain rich and vivid descriptions. Failures happen in general due to factual hallucinations or missed details. \autoref{fig:bad_captioning} shows, for instance, a failing test with a score of 0.56.

\begin{figure}[h]
    \begin{promptbox}
    {\small\ttfamily
   The image captures the clean interior of a modern car, viewed from the rear. A driver, wearing a white shirt and jeans, is buckled into the driver's seat, holding a phone to their ear with their right hand while their left hand rests on the steering wheel. The light gray front passenger and rear seats are unoccupied.
    }
    \end{promptbox}
    \caption{Example of a produced scene description with a low similarity (0.56) to the reference caption. The driver's gender and the seat belt status are missing in the description.}
    \label{fig:bad_captioning}
\end{figure}

\subsection{Expert Feedback}\label{sec:expert-feedback}

We discussed the evaluation results of ISU-Test with a company expert who has more than 20 years of experience in developing vision-based systems such as ISU and was not involved in this work. The discussion was structured around the following questions:

\textit{Based on the results, is ISU-Test helpful for benchmarking ISU systems developed within the company?} ISU-Test shows potential as a tool for evaluating prototypical ISU implementations, particularly for identifying failure cases and performance limitations. It is especially suitable for exploring complex and diverse scene configurations, including variations in in-vehicle objects, driver poses, and vehicle models.

\textit{How realistic are the scenes generated by ISU-Test?} The generated scenes enable the representation of a wide range of conditions. Compared to existing synthetic interior scene datasets and generation approaches, ISU-Test provides more realistic and configurable scene appearances.

\textit{Beyond benchmarking, do you see additional application domains for ISU-Test?} Simulator-based scene generation facilitates also the creation of large static datasets for comparing different ISU implementations. This capability is not limited to benchmarking but can also support training and fine-tuning of ISU systems.
In practice, \toolname can be integrated into existing validation workflows as a complementary component to dataset-based evaluation. For instance, it can be used during development to identify failure-prone regions of the input space, and in later stages to perform regression testing across model updates. Compared to manual data collection, the ability to systematically generate targeted scenarios enables faster iteration and more comprehensive coverage.

\section{Related Work}
\label{sec:related-work}

\subsection{Datasets}

SVIRO~\cite{sviro2024} is a Synthetic dataset for Vehicle Interior Rear seat Occupancy detection as well for classification tasks compromising different data modalities such RGB image, depth image, IR images covering different vehicle types and human appearances. TICAM~\cite{katrolia2021ticam} is an in cabin monitoring dataset compromising synthetic as well as driving simulator recorded and human labeled images focusing on the front interior (more than 100,000). Drive\&Act~\cite{driveandact2019} is a dataset which focuses on static simulator in cabin recordings of fine grained driver activities such as the usage of laptops or in-car entertainment functions. 

However, all these datasets have in common that they can become part of the training data of an VLM-based in cabin monitoring system limiting their usability for failure/inaccuracy detection. At the same, there is no guidance to potential corner cases what is in contrast enabled in our optimization-based approach.

\subsection{Test Generation}

Attaoui and Pastore~\cite{attaoui2025designator} developed DESIGNATOR, a toolset for generating datasets for testing and retraining deep neural networks (DNNs) performing computer vision tasks in Martian-like environments. 
While their approach can also generate synthetic images through simulation and relies on parameter optimization, it is limited to the testing of DNNs focusing solely on the mars environment simulation.
Haq et al.~\cite{Haq_2021} used as ISU-Test an optimization based approach  and a rendering engine to find facial expression/images making a key-point detection systems misclassify key points. Also this approach is limited to DNN testing and uses a specific simulation setup. Baresi et al.~\cite{baresi2025}
apply diffusion-based transformation techniques to augment driving scenarios when testing a vision-based lane-keeping assistant.

To the best of our knowledge, \toolname is the first approach to implement automated and guided test generation for benchmarking VLM-based in-car scene understanding systems.
\section{Conclusions and Future Work}\label{sec:conclusion}

We presented \toolname, an automated testing framework for evaluating vision-language model (VLM)-based in-car scene understanding systems. Our approach combines controllable, rendering-based scene generation with search-based testing to systematically explore the space of in-cabin scenarios and identify failure-inducing inputs.
Our evaluation on both open-source and industrial systems demonstrates that \toolname consistently outperforms randomized testing, achieving substantially higher failure rates and improved failure coverage while maintaining comparable or higher diversity. Importantly, the agreement between simulated and real-world evaluations indicates that failures discovered in simulation largely transfer to physical settings, supporting the use of simulation-based testing as a practical proxy for early-stage validation.
These findings highlight a key limitation of current validation practices for VLM-based systems: static datasets are insufficient to expose rare, safety-critical behaviors. Instead, systematic and controllable test generation is required to ensure robustness under diverse and previously unseen conditions.

Future work will extend \toolname along several dimensions. First, we plan to integrate behaviorally consistent generative models (e.g., diffusion- or control-based approaches) to improve photorealism while preserving semantic correctness. Second, we aim to support temporal and multi-view reasoning to evaluate dynamic scenarios involving driver actions and object interactions over time to detect safety-critical situations.

\section{Data Availability Statement}

We provide a replication package with our framework, evaluation scripts, and datasets~\cite{repo} for the studies with standalone VLMs for in-car scene understanding.
%
%
The SUT and data related to the industrial case study cannot be shared due to confidentiality constraints; for these, we report only aggregate results.


\bibliographystyle{IEEEtran}
\bibliography{paper}

@inproceedings{SMPL-X:2019,
	title        = {Expressive Body Capture: {3D} Hands, Face, and Body from a Single Image},
	author       = {Pavlakos, Georgios and Choutas, Vasileios and Ghorbani, Nima and Bolkart, Timo and Osman, Ahmed A. A. and Tzionas, Dimitrios and Black, Michael J.},
	year         = 2019,
	booktitle    = {Proceedings IEEE Conf. on Computer Vision and Pattern Recognition (CVPR)},
	pages        = {10975--10985}
}

@article{2026-Guo-OJ-ITS,
  title = {Foundation Models in Autonomous Driving: A Survey on Scenario Generation and Scenario Analysis},
  author = {Gao, Yuan and Piccinini, Mattia and Zhang, Yuchen and Wang, Dingrui and Moller, Korbinian and Brusnicki, Roberto and Zarrouki, Baha and Gambi, Alessio and Totz, Jan Frederik and Storms, Kai and Peters, Steven and Stocco, Andrea and Alrifaee, Bassam and Pavone, Marco and Betz, Johannes},
  year = {2026},
  journal = {IEEE Open Journal of Intelligent Transportation Systems},
  publisher = {IEEE},
  url = {https://arxiv.org/abs/2506.11526},
}

@article{RiccioEMSE20,
	title        = {Testing machine learning based systems: a systematic mapping},
	author       = {Riccio, Vincenzo and Jahangirova, Gunel and Stocco, Andrea and Humbatova, Nargiz and Weiss, Michael and Tonella, Paolo},
	year         = 2020,
	journal      = {Empirical Software Engineering},
	volume       = 25
}

@article{Wilcoxon1945,
	title        = {Individual Comparisons by Ranking Methods},
	author       = {Frank Wilcoxon},
	year         = 1945,
	journal      = {Biometrics Bulletin},
	volume       = 1,
	number       = 6
}

@inproceedings{humbatova2020taxonomy,
	title        = {Taxonomy of real faults in deep learning systems},
	author       = {Humbatova, Nargiz and Jahangirova, Gunel and Bavota, Gabriele and Riccio, Vincenzo and Stocco, Andrea and Tonella, Paolo},
	year         = 2020,
	booktitle    = {Proceedings of the ACM/IEEE 42nd international conference on software engineering}
}

@inproceedings{2024-Lambertenghi-ICST,
	title        = {Assessing Quality Metrics for Neural Reality Gap Input Mitigation in Autonomous Driving Testing},
	author       = {Stefano Carlo Lambertenghi and Andrea Stocco},
	year         = 2024,
	booktitle    = {Proceedings of 17th IEEE International Conference on Software Testing, Verification and Validation},
	series       = {ICST '24}
}

@misc{repo,
	title        = {Replication Package},
	author       = {Anonymous},
	howpublished = {\url{https://figshare.com/s/cb5b0eae0411e54b1bbd}}
}

@artifactsoftware{R,
	title        = {R: A Language and Environment for Statistical Computing},
	author       = {{R Core Team}},
	year         = 2019,
	organization = {R Foundation for Statistical Computing}
}

@article{surrogate2024biagiola,
	title        = {Testing of Deep Reinforcement Learning Agents with Surrogate Models},
	author       = {Biagiola, Matteo and Tonella, Paolo},
	year         = 2024,
	journal      = {ACM Trans. Softw. Eng. Methodol.},
	volume       = 33,
	number       = 3,
	issn         = {1049-331X},
	issue_date   = {March 2024},
	articleno    = 73,
	numpages     = 33
}

@inproceedings{feldt2016diversity,
	title        = {Test Set Diameter: Quantifying the Diversity of Sets of Test Cases},
	author       = {Feldt, Robert and Poulding, Simon and Clark, David and Yoo, Shin},
	year         = 2016,
	booktitle    = {2016 IEEE International Conference on Software Testing, Verification and Validation (ICST)},
	volume       = {},
	number       = {}
}

@article{mishraIncabin2022,
	title        = {In-Cabin Monitoring System for Autonomous Vehicles},
	author       = {Mishra, Ashutosh and Lee, Sangho and Kim, Dohyun and Kim, Shiho},
	year         = 2022,
	journal      = {Sensors},
	volume       = 22,
	number       = 12,
	doi          = {10.3390/s22124360},
	issn         = {1424-8220},
	url          = {https://www.mdpi.com/1424-8220/22/12/4360},
	article-number = 4360,
	pubmedid     = 35746138
}

@techreport{EuroNCAP2025_AssistedDriving,
	title        = {Assessment Protocol -- Assisted Driving: Highways \& Interurban Assist Systems. Technical Bulletin SD 202 -- Driver Monitoring Test Procedure},
	author       = {European New Car Assessment Programme (Euro NCAP)},
	year         = 2025,
	month        = mar,
	url          = {https://www.euroncap.com/media/85831/euro-ncap-protocol-assisted-driving-v10.pdf},
	note         = {Implementation January 2026},
	institution  = {Euro NCAP},
	version      = {1.0}
}

@techreport{Diederichs2025_InCabinMonitoring,
	title        = {Pioneering In-Cabin Monitoring: Unmasking the Power of 2D and 3D Cameras through Sensor Fusion},
	author       = {Frederik Diederichs and Fraunhofer IOSB},
	year         = 2025,
	month        = {—},
	url          = {https://www.iosb.fraunhofer.de/content/dam/iosb/iosbtest/documents/kompetenzen/bildauswertung/hai/projekte-und-produkte/White%20Paper_In-Cabin%20Monitoring_2025_red..pdf},
	note         = {White Paper},
	institution  = {Fraunhofer Institute of Optronics, System Technologies and Image Exploitation (IOSB)}
}

@inproceedings{Haq_2021,
	title        = {Automatic test suite generation for key-points detection DNNs using many-objective search (experience paper)},
	author       = {Haq, Fitash Ul and Shin, Donghwan and Briand, Lionel C. and Stifter, Thomas and Wang, Jun},
	year         = 2021,
	month        = jul,
	booktitle    = {Proceedings of the 30th ACM SIGSOFT International Symposium on Software Testing and Analysis},
	publisher    = {ACM},
	series       = {ISSTA ’21},
	pages        = {91–102},
	doi          = {10.1145/3460319.3464802},
	url          = {http://dx.doi.org/10.1145/3460319.3464802},
	collection   = {ISSTA ’21}
}

@misc{EuroNCAP2025,
	title        = {The European New Car Assessment Programme},
	author       = {{Euro NCAP}},
	year         = 2025,
	note         = {Accessed: 2025-10-26},
	howpublished = {\url{https://www.euroncap.com/en}}
}

@inproceedings{baresi2025,
	title        = {Efficient Domain Augmentation for Autonomous Driving Testing Using Diffusion Models},
	author       = {Baresi, Luciano and Xian Hu, Davide Yi and Stocco, Andrea and Tonella, Paolo},
	year         = 2025,
	booktitle    = {2025 IEEE/ACM 47th International Conference on Software Engineering (ICSE)},
	volume       = {},
	number       = {},
	pages        = {398--410},
	doi          = {10.1109/ICSE55347.2025.00206}
}

@article{attaoui2025designator,
	title        = {Search-Based DNN Testing and Retraining With GAN-Enhanced Simulations},
	author       = {Attaoui, Mohammed Oualid and Pastore, Fabrizio and Briand, Lionel C.},
	year         = 2025,
	month        = apr,
	journal      = {IEEE Trans. Softw. Eng.},
	publisher    = {IEEE Press},
	volume       = 51,
	number       = 4,
	pages        = {1086–1103},
	doi          = {10.1109/TSE.2025.3540549},
	issn         = {0098-5589},
	url          = {https://doi.org/10.1109/TSE.2025.3540549},
	issue_date   = {April 2025},
	numpages     = 18
}

@inproceedings{sviro2024,
	title        = {SVIRO: Synthetic Vehicle Interior Rear Seat Occupancy},
	author       = {Steve Dias Da Cruz and Oliver Wasenm¨uller and Hans-Peter Beise and Thomas Stifter and Didier Stricker},
	year         = 2024,
	month        = {dec},
	publisher    = {TIB},
	doi          = {10.57702/n949p0zq},
	url          = {https://service.tib.eu/ldmservice/dataset/sviro--synthetic-vehicle-interior-rear-seat-occupancy},
	institution  = {No Organization}
}

@inproceedings{katrolia2021ticam,
	title        = {TICaM: {A} Time-of-flight In-car Cabin Monitoring Dataset},
	author       = {Jigyasa Singh Katrolia and Ahmed El{-}Sherif and Hartmut Feld and Bruno Mirbach and Jason R. Rambach and Didier Stricker},
	year         = 2021,
	booktitle    = {32nd British Machine Vision Conference 2021, {BMVC} 2021, Online, November 22-25, 2021},
	publisher    = {{BMVA} Press},
	pages        = 277,
	url          = {https://www.bmvc2021-virtualconference.com/assets/papers/0701.pdf},
	bibsource    = {dblp computer science bibliography, https://dblp.org}
}

@inproceedings{driveandact2019,
	title        = {Drive\&Act: A Multi-Modal Dataset for Fine-Grained Driver Behavior Recognition in Autonomous Vehicles},
	author       = {Martin, Manuel and Roitberg, Alina and Haurilet, Monica and Horne, Matthias and Rei{\ss}, Simon and Voit, Michael and Stiefelhagen, Rainer},
	year         = 2019,
	booktitle    = {2019 IEEE/CVF International Conference on Computer Vision (ICCV)},
	volume       = {},
	number       = {},
	pages        = {2801--2810},
	doi          = {10.1109/ICCV.2019.00289}
}

@article{zhang2019bertscore,
	title        = {Bertscore: Evaluating text generation with bert},
	author       = {Zhang, Tianyi and Kishore, Varsha and Wu, Felix and Weinberger, Kilian Q and Artzi, Yoav},
	year         = 2019,
	journal      = {arXiv preprint arXiv:1904.09675}
}

@inproceedings{meteor,
	title        = {Meteor: an automatic metric for MT evaluation with high levels of correlation with human judgments},
	author       = {Lavie, Alon and Agarwal, Abhaya},
	year         = 2007,
	booktitle    = {Proceedings of the Second Workshop on Statistical Machine Translation},
	location     = {Prague, Czech Republic},
	publisher    = {Association for Computational Linguistics},
	address      = {USA},
	series       = {StatMT '07},
	pages        = {228–231},
	numpages     = 4
}

@inproceedings{bleu,
	title        = {BLEU: a method for automatic evaluation of machine translation},
	author       = {Papineni, Kishore and Roukos, Salim and Ward, Todd and Zhu, Wei-Jing},
	year         = 2002,
	booktitle    = {Proceedings of the 40th Annual Meeting on Association for Computational Linguistics},
	location     = {Philadelphia, Pennsylvania},
	publisher    = {Association for Computational Linguistics},
	address      = {USA},
	series       = {ACL '02},
	pages        = {311–318},
	doi          = {10.3115/1073083.1073135},
	numpages     = 8
}

@misc{karpathy2015deepvisualsemanticalignmentsgenerating,
	title        = {Deep Visual-Semantic Alignments for Generating Image Descriptions},
	author       = {Andrej Karpathy and Li Fei-Fei},
	year         = 2015,
	url          = {https://arxiv.org/abs/1412.2306},
	eprint       = {1412.2306},
	archiveprefix = {arXiv},
	primaryclass = {cs.CV}
}

@misc{vinyals2015tellneuralimagecaption,
	title        = {Show and Tell: A Neural Image Caption Generator},
	author       = {Oriol Vinyals and Alexander Toshev and Samy Bengio and Dumitru Erhan},
	year         = 2015,
	url          = {https://arxiv.org/abs/1411.4555},
	eprint       = {1411.4555},
	archiveprefix = {arXiv},
	primaryclass = {cs.CV}
}

@misc{agrawal2016vqavisualquestionanswering,
	title        = {VQA: Visual Question Answering},
	author       = {Aishwarya Agrawal and Jiasen Lu and Stanislaw Antol and Margaret Mitchell and C. Lawrence Zitnick and Dhruv Batra and Devi Parikh},
	year         = 2016,
	url          = {https://arxiv.org/abs/1505.00468},
	eprint       = {1505.00468},
	archiveprefix = {arXiv},
	primaryclass = {cs.CL}
}

@misc{bernardi2017automaticdescriptiongenerationimages,
	title        = {Automatic Description Generation from Images: A Survey of Models, Datasets, and Evaluation Measures},
	author       = {Raffaella Bernardi and Ruket Cakici and Desmond Elliott and Aykut Erdem and Erkut Erdem and Nazli Ikizler-Cinbis and Frank Keller and Adrian Muscat and Barbara Plank},
	year         = 2017,
	url          = {https://arxiv.org/abs/1601.03896},
	eprint       = {1601.03896},
	archiveprefix = {arXiv},
	primaryclass = {cs.CL}
}

@misc{vedantam2015ciderconsensusbasedimagedescription,
	title        = {CIDEr: Consensus-based Image Description Evaluation},
	author       = {Ramakrishna Vedantam and C. Lawrence Zitnick and Devi Parikh},
	year         = 2015,
	url          = {https://arxiv.org/abs/1411.5726},
	eprint       = {1411.5726},
	archiveprefix = {arXiv},
	primaryclass = {cs.CV}
}

@misc{kilickaya2016reevaluatingautomaticmetricsimage,
	title        = {Re-evaluating Automatic Metrics for Image Captioning},
	author       = {Mert Kilickaya and Aykut Erdem and Nazli Ikizler-Cinbis and Erkut Erdem},
	year         = 2016,
	url          = {https://arxiv.org/abs/1612.07600},
	eprint       = {1612.07600},
	archiveprefix = {arXiv},
	primaryclass = {cs.CL}
}

@article{gatys2016nst,
	title        = {Image style transfer using convolutional neural networks},
	author       = {Gatys, Leon A and Ecker, Alexander S and Bethge, Matthias},
	year         = 2016,
	journal      = {Journal of Vision},
	volume       = 16,
	number       = 12,
	pages        = 326,
	doi          = {10.1167/16.12.326}
}

@article{alimisis2025advances,
	title        = {Advances in Diffusion Models for Image Data Augmentation: A Review of Methods, Models, Evaluation Metrics and Future Research Directions},
	author       = {Panagiotis Alimisis and Ioannis Mademlis and Panagiotis Radoglou-Grammatikis and Panagiotis Sarigiannidis and Georgios Th. Papadopoulos},
	year         = 2025,
	journal      = {Artificial Intelligence Review},
	volume       = 58,
	pages        = 112,
	doi          = {10.1007/s10462-025-11116-x},
	url          = {https://doi.org/10.1007/s10462-025-11116-x}
}

@inproceedings{sorokin2026stellar,
	title        = {{STELLAR}: A Search-Based Testing Framework for Large Language Model Applications},
	author       = {Sorokin, Lev and Vasilev, Ivan and Friedl, Ken E. and Stocco, Andrea},
	year         = 2026,
	booktitle    = {Proceedings of the 33rd IEEE International Conference on Software Analysis, Evolution and Reengineering},
	publisher    = {IEEE}
}

@misc{nanobana2026,
	title        = {Nano Banana AI Image Generator},
	author       = {{Nanobana}},
	year         = 2026,
	note         = {AI‑based image generation and editing platform},
	howpublished = {\url{https://www.nanobana.net/}}
}

@misc{nvidiacosmos2026,
	title        = {NVIDIA Cosmos: World Foundation Models for Physical AI},
	author       = {{NVIDIA Corporation}},
	year         = 2026,
	note         = {Open platform with world foundation models and data processing for robotics, autonomous systems, and physical‑AI research},
	howpublished = {\url{https://www.nvidia.com/en-us/ai/cosmos/}}
}

@article{hamming1950error,
  author  = {Hamming, Richard W.},
  title   = {Error Detecting and Error Correcting Codes},
  journal = {Bell System Technical Journal},
  volume  = {29},
  number  = {2},
  pages   = {147--160},
  year    = {1950}
}

@article{Vargha-Delaney,
 ISSN = {10769986, 19351054},
 URL = {http://www.jstor.org/stable/1165329},
 author = {András Vargha and Harold D. Delaney},
 journal = {Journal of Educational and Behavioral Statistics},
 number = {2},
 pages = {101--132},
 publisher = {[American Educational Research Association, Sage Publications, Inc., American Statistical Association]},
 title = {A Critique and Improvement of the "CL" Common Language Effect Size Statistics of McGraw and Wong},
 urldate = {2026-04-30},
 volume = {25},
 year = {2000}
}

@manual{blender_shrinkwrap_36,
  title        = {Shrinkwrap Modifier},
  author       = {{Blender Foundation}},
  organization = {Blender Foundation},
  year         = {2023},
  url          = {https://docs.blender.org/manual/id/3.6/modeling/modifiers/deform/shrinkwrap.html},
  note         = {Blender 3.6 Manual, accessed 2026-04-23}
}

@misc{EU2019_GSR,
  title = {Regulation (EU) 2019/2144 on type-approval requirements for motor vehicles},
  author = {{European Parliament and Council}},
  year = {2019},
  note = {Official Journal of the European Union}
}

\end{document}